\pdfoutput=1

\documentclass[11pt]{article}

\usepackage[]{ACL2023}

\usepackage{times}
\usepackage{latexsym}
\usepackage{helvet} 
\usepackage{courier}
\usepackage{graphicx}
\usepackage{xcolor}
\usepackage{soul} 
\usepackage{amssymb}
\usepackage{amsmath}
\usepackage{pifont}
\usepackage{enumitem}
\usepackage{booktabs}
\usepackage{multirow}
\usepackage{colortbl}
\usepackage{array}
\usepackage{tabularx}
\usepackage{pifont}
\usepackage{bm}
\usepackage{tabularray}
\usepackage{makecell}
\usepackage{stackengine}
\usepackage{tcolorbox}
\tcbuselibrary{breakable}
\newsavebox{\measurebox}

\usepackage[T1]{fontenc}
\usepackage[utf8]{inputenc}
\usepackage{microtype}
\usepackage{inconsolata}
\usepackage{subfig}

\definecolor{myred}{RGB}{255,200,200}
\definecolor{myblue}{RGB}{200,220,255}
\definecolor{mygray}{RGB}{190,190,190}
\definecolor{method}{RGB}{217,217,252}

\newcommand{\hlmethod}[1]{\sethlcolor{method}\hl{#1}}

\newcolumntype{L}{>{\raggedright\arraybackslash}X}

\newcolumntype{T}{>{\fontsize{9.5}{10}\selectfont}c}
\newcolumntype{H}{>{\setbox0=\hbox\bgroup}c<{\egroup}@{}}

\title{Leaps Beyond the Seen: Reinforced Reasoning Augmented Generation for Clinical Notes}

\author{
\textbf{Lo Pang-Yun Ting$^{\text{\ding{171}}\dagger}$},
\textbf{Chengshuai Zhao$^{\text{\ding{170}}\dagger}$}, 
\textbf{Yu-Hua Zeng$^{\text{\ding{171}}}$},
\textbf{Yuan Jee Lim$^{\text{\ding{171}}}$},\\
\textbf{Kun-Ta Chuang$^{\text{\ding{171}}}$}, 
\textbf{Huan Liu$^{\text{\ding{170}}}$}\\ 
$^{\text{\ding{171}}}$Dept. of Computer Science and Information Engineering, National Cheng Kung University \\ $^{\text{\ding{170}}}$School of Computing and Augmented Intelligence, Arizona State University\\ 
\texttt{\{lpyting, yhzeng, yjlim\}@netdb.csie.ncku.edu.tw}, \\
\texttt{ktchuang@mail.ncku.edu.tw}, \texttt{\{czhao93, huanliu\}@asu.edu}\\
}

\begin{document}
\maketitle
\def\thefootnote{$\dagger$}\footnotetext{Equal contribution.}

\begin{abstract}

Clinical note generation aims to produce free-text summaries of a patient's condition and diagnostic process, with discharge instructions being a representative long-form example. While recent LLM-based methods pre-trained on general clinical corpora show promise in clinical text generation, they fall short in producing long-form notes from limited patient information. In this paper, we propose \textbf{\emph{ReinRAG}}, a reinforced reasoning augmented generation (RAG) for long-form discharge instructions based on pre-admission information. \emph{ReinRAG} retrieves reasoning paths from a medical knowledge graph to provide explicit semantic guidance to the LLM. To bridge the information gap, we propose group-based retriever optimization (GRO) which improves retrieval quality with group-normalized rewards, encouraging reasoning leaps for deeper inference by the LLM. Comprehensive experiments on the real-world dataset show that \emph{ReinRAG} outperforms baselines in both clinical efficacy and natural language generation metrics. Further analysis reveals that \emph{ReinRAG} fills semantic gaps in sparse input scenarios, and retrieved reasoning paths help LLMs avoid clinical misinterpretation by focusing on key evidence and following coherent reasoning.

\end{abstract}
\section{Introduction}

Clinical note generation improves communications and decision-making among healthcare professionals and patients, while also reducing the time burden of manually writing reports~\cite{arndt2017tethered, sinsky2016allocation}. This has motivated research into using large language models (LLMs) for automatic clinical note and report generation~\cite{abacha2023empirical, jin2024promptmrg, liu2024bootstrapping}. Nevertheless, most works focus on generating short or structured summaries that address specific elements, such as diagnoses or treatments, instead of producing extensive and in-depth outputs.

\begin{figure}[t]
\graphicspath{{figs/}}
\begin{center}
\includegraphics[width=0.5\textwidth]{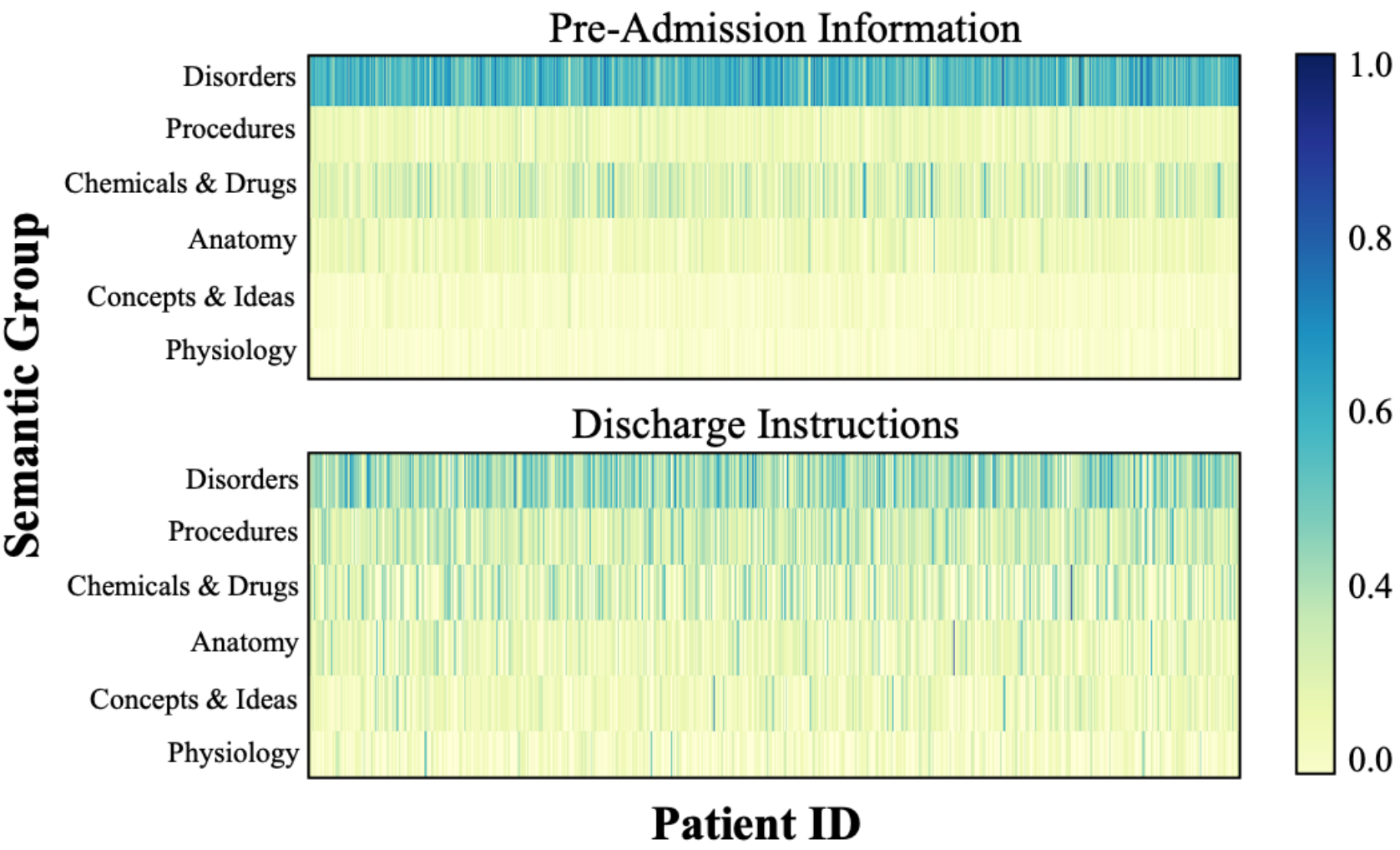}
\end{center}
\caption{Keyword distribution across UMLS semantic clusters in patients' pre-admission information and discharge instructions. Keywords from pre-admission information are concentrated in the \textit{Disorders} cluster, whereas those in discharge instructions span a broader range of semantic clusters, revealing a substantial information gap.}
\label{fig:data_analysis}
\end{figure}

\textit{Patient discharge instruction} summarizes a wide range of information, including diagnoses, medications, and the patient’s condition during hospitalization~\cite{kononenko2001machine,ting2025cand,ting2025early,kononenko2001machine}, while also providing guidance for post-discharge care~\cite{kind2011documentation, gonccalves2016discharge}. Automatically generating discharge instructions can reduce the workload for clinicians. Moreover, generating preliminary discharge instructions could provide clinicians with an early snapshot of likely diagnoses, treatments, and follow-up needs, serving as a useful reference throughout the hospital stay. Despite its significance, \textit{the automatic generation of discharge instructions from pre-admission information} remains largely underexplored and faces following challenges.

\textbf{Challenge 1: Open-ended generation without explicit evidence.} Generating discharge instruction is inherently an open-ended generation task, where the correct content may not be explicitly present in the data. Most medical LLMs~\cite{qiu2024towards, wu2024pmc} or Retrieval-Augmented Generation (RAG)~\cite{lewis2020retrieval} models~\cite{zakka2024almanac, lozano2023clinfo, xiong2024benchmarking} are pre-trained on general clinical corpora and are mainly designed for answering questions with explicit evidence or solving closed-ended tasks with predefined answer choices. As a result, they may not be well suited to our scenario. 

\textbf{Challenge 2: Information gap in discharge instructions.} There is a significant information gap between patients' pre-admission data and discharge instructions, as the latter typically relies on hospital-stay information. Without proper guidance, LLMs only generate semantically similar content from pre-admission inputs and fail to infer deeper clinical states.

To analyze information gap in the second challenge, we select 500 de-identified patients' discharge summaries from MIMIC-IV-note~\cite{johnson2023mimic, goldberger2000components}, which contains data from the Beth Israel Deaconess Medical Center. For each patient, we define pre-admission information as allergies, chief complaints, and the history of present illness (HPI), and compare it with their discharge instructions. We then extract keywords from each text and map them to semantic clusters\footnote{In this paper, semantic clusters refer to the semantic groups defined in the UMLS semantic network.} in the UMLS (Unified Medical Language System)~\cite{bodenreider2004unified, UMLS2024AB}, which is a comprehensive medical knowledge base structured as a large-scale knowledge graph (KG). Figure~\ref{fig:data_analysis} presents the distribution of extracted keywords across UMLS semantic clusters, revealing the substantial content difference between the patients' pre-admission information and their discharge instructions. This indicates that, LLMs need to be guided on when to perform fine-grained reasoning~\cite{zhang2025ratt,zhao2025chain,liu2024much} to infer more details from known situations (e.g., patient symptoms), and when to perform jump thinking to infer deeper information (e.g., diagnoses or treatments) to bridge the information gap.

These challenges suggest that generating accurate instructions involves two key components: \textbf{retrieving external knowledge to provide reasoning direction} that guides accurate long-form generation, and \textbf{controlling the granularity of reasoning steps} to help LLMs infer possible downstream clinical details beyond the observed input. In response, we propose the \textbf{\emph{ReinRAG}} model (\underline{\textbf{Rein}}forced \underline{\textbf{R}}easoning \underline{\textbf{A}}ugmentation for Clinical Note \underline{\textbf{G}}eneration) for long-form discharge instruction generation based on pre-admission information. \textbf{To retrieve useful knowledge and ensure accurate reasoning direction}, we incorporate the UMLS KG to retrieve structured reasoning paths, providing LLMs with explicit semantic guidance in open-ended generation. \textbf{To control the LLM’s reasoning granularity}, we design a retriever based on reinforcement learning (RL) that learns to select reasoning paths exhibiting \underline{reasoning leaps} across semantic clusters in the KG. Unlike conventional RAG approaches that rely on single-hop or simple multi-hop retrieval, our method uses RL to optimize the retrieval and guide LLMs on when to retrieve semantically similar concepts or make reasoning leaps to obtain more diverse information. This design helps the LLM advance its reasoning and bridge the information gap when only pre-admission information is available. Furthermore, inspired by Group Relative Policy Optimization (GRPO)~\cite{shao2024deepseekmath}, we proposed a novel optimization mechanism, named \emph{GRO} (\underline{G}roup-Based \underline{R}etriever \underline{O}ptimization), which retrieves multiple reasoning paths per patient and assigns group-normalized rewards to discover the most informative semantic paths. Our key contributions are summarized as follows:

\begin{itemize}[leftmargin=*]
    \item [$\star$]\textbf{Discharge Instruction Generation with Limited Information.} We target the challenging task of generating long-form discharge instructions using only patients' pre-admission data, going beyond conventional short-form generation. This represents a \textit{new and largely unexplored direction with potential clinical value in early decision support}.
    \item [$\star$]\textbf{Reinforced Reasoning Augmentation.} We enhance RAG with a novel RL-based retriever that performs \textit{reasoning leaps} across semantic clusters in a medical KG. This guides the LLM to bridge the gap between limited pre-admission inputs and complex discharge instructions, marking a \textit{pioneering application of RL for reasoning-based retrieval in long-form generation}.
    \item [$\star$]\textbf{Group-Based Retriever Optimization.} We introduce \emph{GRO}, a novel RL optimization strategy that retrieves multiple reasoning paths per input and leverages group-normalized rewards to effectively guide LLM generation.
    \item [$\star$]\textbf{Practical Effectiveness.} Experiments on the real-world MIMIC-IV-note dataset demonstrate that \emph{ReinRAG} consistently outperform baselines in both clinical efficacy and natural language generation, producing more accurate and less irrelevant information.
\end{itemize}

\section{Related Work}

\subsection{Medical-Specialized LLMs}
A growing number of medical-specialized LLMs have been pre-trained on clinical corpora, including Meditron~\cite{chen2023meditron}, ClinicalGPT~\cite{wang2023clinicalgpt}, HuatuoGPT~\cite{zhang2023huatuogpt}, PediatricsGPT~\cite{yang2024pediatricsgpt}, ClinicalMamba~\cite{yang2024clinicalmamba}, BioMistral~\cite{DBLP:conf/acl/LabrakBMGRD24}, PMC-LLaMA~\cite{wu2024pmc}, and MMed-Llama3~\cite{qiu2024towards}. These models improve fluency and factuality on tasks such as ICD coding and short-form clinical QA.

\subsection{Retrieval-Augmented Generation in Medical}
Retrieval-augmented generation (RAG) techniques play a predominant role in the medical domain by enhancing clinical text generation~\cite{li2023chatdoctor, zakka2024almanac, lozano2023clinfo, xiong2024benchmarking, xiong2024improving, wu2024medical}. Recent studies have incorporated knowledge graph (KG) retrieval to guide LLMs toward concise clinical answers. For instance, MindMap~\cite{wen2023mindmap}, Knowledge Seeds~\cite{wu2024guiding}, and DR.KNOWS~\cite{gao2025leveraging} retrieve relevant KG triples or paths to prompt the model.
Most focus on questions that have direct answers in a single document or involve selecting limited answer options and short-form outputs such as diagnostic options, probable diseases, or drug recommendations.

\subsection{Clinical Note Generation}
Other efforts focus on distinct settings, such as summarizing doctor–patient dialogues~\cite{abacha2023empirical} or generating radiology reports~\cite{jin2024promptmrg, liu2024bootstrapping, yin2019automatic} from X-ray images.

Although the above approaches achieve strong performance within their respective settings, they mainly focus on generating short-form outputs. A few studies have explored the generation of long-form discharge summaries~\cite{li2024llamacare, wu2024epfl, williams2024evaluating, ellershaw2024automated}, but these efforts typically rely on rich in-hospital data, such as progress notes or complete EHRs, that only become available after a prolonged hospital stay. By contrast, we tackle a more challenging scenario of generating long-form discharge instructions using only pre-admission data and design an RL-based retriever over the medical knowledge graph to augment LLM generation.

\section{Methodology}

\begin{figure*}[t]
\graphicspath{{figs/}}
\begin{center}
\includegraphics[width=1.0\textwidth]{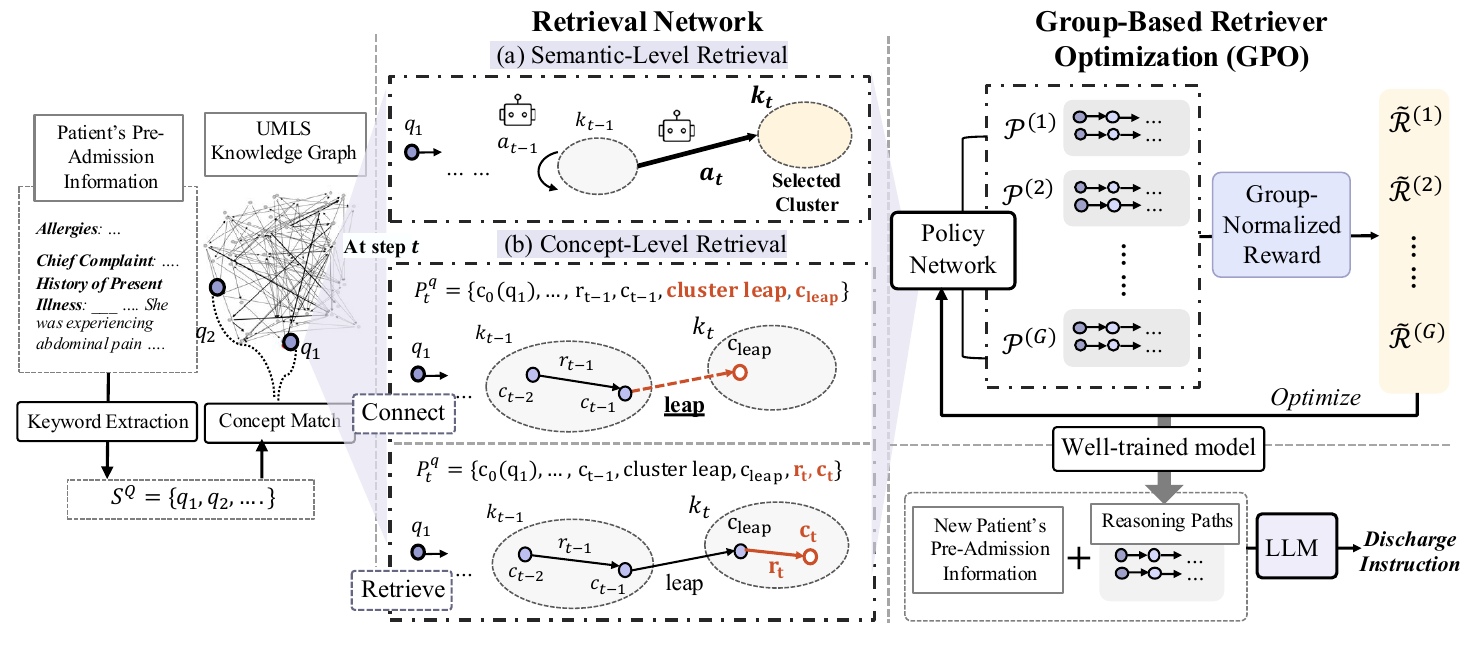}
\end{center}
\caption{The overview of \emph{ReinRAG}. After extracting the patient's pre-admission information and matching keywords with the UMLS KG, the retrieval network performs two-level retrieval based on RL to form reasoning paths. Then, the group-based retriever optimization leverages group-relative rewards to optimize the policy network. Finally, the well-trained \emph{ReinRAG} generates reasoning paths to guide the LLM in discharge instruction generation.}
\label{fig:framework}
\end{figure*}

The proposed model \underline{\textbf{Rein}}forced \underline{\textbf{R}}easoning \underline{\textbf{A}}ugmentation for Clinical Note \underline{\textbf{G}}eneration, \textbf{\emph{ReinRAG}}, consists of two main components, as illustrated in Figure~\ref{fig:framework}: 
(1) \textit{Retrieval Network} (Sec.~\ref{subsec:retrieval_network}), which controls reasoning granularity by performing two-level of retrievals based on RL; and (2) \textit{Group-Based Retriever Optimization} (Sec.~\ref{subsec:gro}), which optimizes the model based on a group of reasoning paths to guide long-form discharge instruction generation.

\subsection{Basic Setup}

\subsubsection{\textbf{Notations and Problem Definition.}} Our goal is to retrieve reasoning paths from a medical KG to guide LLM generation. Formally, a \textbf{medical knowledge graph} (UMLS KG~\cite{bodenreider2004unified, UMLS2024AB} used in our paper) is represented as $\mathcal{G} = \{(c, r, c') \mid c, c' \in C, r \in R\}$, where $C$ is the set of medical concepts and $R$ is the set of relations. A triplet $(c, r, c')$ describes the relationship between two concepts, such as (``\textit{dyspnea care}'', ``\textit{focus of}'', ``\textit{breathlessness care management}''). Let $\mathcal{G}^k$ denote the set of semantic clusters, where each concept $c \in \mathcal{C}$ belongs to a specific cluster $k \in \mathcal{G}^k$ based on its semantic (e.g., concept ``\textit{dyspnea care}'' belongs to cluster ``\textit{Procedures}''). For \textbf{patient information}, let $Q$ be the pre-admission information and $S^Q$ be the set of keywords extracted from $Q$. Each keyword $q\in S^Q$ can be mapped to a specific concept $c$ in KG $\mathcal{G}$.\footnote{We describe  the extracted terms as ``keywords'' and KG nodes as ``concepts'' for clarity.}

Each \textbf{reasoning path} starts from a keyword $q\in S^Q$ and is denoted as $P^q_t$ at retrieval step $t$, with $P^q_0 = \{q\}$. Therefore, given $Q$, $\mathcal{G}$, and initial reasoning paths $\{P^q_0\}_{q \in S^Q}$, we aim to retrieve and extend reasoning paths to guide LLM generation.

\subsubsection{\textbf{Retrieval Environment Formulation.}}
Our task is viewed as a Markov Decision Process (MDP), where the retriever decides whether to continue exploring concepts within the current cluster or to leap to another cluster.

\noindent\textbf{State.} The state $(s^k_t, s^c_t) \in \mathcal{S}$ represents the current retrieval situation, consisting of the \textit{cluster state} $s^k_t$ and the \textit{concept state} $s^c_t$, described as follows.

\noindent\textit{\underline{Cluster State $s^k_t$.}} 
The cluster state representation is constructed based on both the currently selected cluster $k_t$ and a scarce cluster $k_{\text{scarce}}$, which is defined as the cluster with the fewest keywords in $S^Q$. This design encourages the retriever to reason not only within the current cluster but also toward underrepresented semantic. The representation $\mathbf{s}^k_t$ of the cluster state is formulated as:

\begin{equation}
\label{eq:state_cluster_embed}
\mathbf{s}^k_t = [\mathbf{k}_t \,\|\, \mathbf{k}_{\text{scarce}}],
\end{equation}
where $\mathbf{k}_t\in \mathbb{R}^{2d}$ and  $\mathbf{k}_{\text{scarce}}\in \mathbb{R}^{2d}$ denote the hidden state embeddings of $k_t$ and $k_{\text{scarce}}$, respectively. The symbol $\|$ represents embedding concatenation.

\noindent\textit{\underline{Concept State $s^c_t$.}} The concept state representation is formulated based on all explored concepts, denoted as $C_t$, as follows:

\begin{equation}
\label{eq:state_concept_embed}
\mathbf{s}^c_t = \mathbf{M}\cdot\text{avg}(\{\mathbf{c} = \text{encoder}(c)|c \in C_t\}),
\end{equation}
where each concept $c$ is encoded using a pretrained SapBERT encoder~\citep{liu2021sapbert}, which is trained on the UMLS dataset. The matrix $\mathbf{M} \in \mathbb{R}^{d \times d}$ is a learnable projection.

\noindent\textbf{Action.} The set of possible actions $A_t\in \mathcal{A}$ at each step $t$ represents  ``leaps'' to another (or the same) clusters in $\mathcal{G}^k$. Formally, an action at step $t$ is defined as $a_t = (k_{t-1} \rightarrow  k_t) \in A_t$, indicating the retriever transitions from cluster $k_{t-1}$ to $k_t$. Each action is represented by the embeddings of the previously visited and the selected clusters, formulated as follows:

\begin{equation}
\label{eq:action_embed}
\mathbf{a}_t = [\mathbf{k}_{t-1}\|\mathbf{k}_{t}].
\end{equation}

After selecting an action, a state transition occurs. The transition function $\delta:\mathcal{S} \times \mathcal{A} \rightarrow \mathcal{S}$ is defined as $\delta((s^k_t, s^c_t), a_t)$, which produces the new state information. Note that at each step, the retriever is allowed to stay in the current cluster or leap to other clusters for the future retrieval. Details of the reward design will be presented in the subsequent sections.

\subsection{Retrieval Network}
\label{subsec:retrieval_network}

Our retriever aims to retrieve reasoning paths from KG $\mathcal{G}$ by controlling reasoning granularity, which involves deciding when to apply reasoning leaps across semantic clusters (semantic-level) and when to select semantically similar concepts in the current cluster (concept-level), forming the two levels of retrieval process, as shown in Figure~\ref{fig:framework}.

\noindent\textbf{Semantic-Level Retrieval.} Following the RL paradigm, our retrieval process is guided by a policy network $\pi_{\theta}$, which determines \textit{which semantic cluster to visit next} based on the current state information ($s^k_t$, $s^c_t$), as show in Figure~\ref{fig:framework}(a)). To align state and action embeddings so that the policy $\pi_{\theta}$ can effectively score their semantic compatibility in a shared representation space, we first map the concatenated state representations $[\mathbf{s}^k_t||\mathbf{s}^c_t]$ through a two-layer feedforward network to obtain a hidden representation $\mathbf{z}_t$. Based on $\mathbf{z}_t$, the policy distribution $\mathbf{d}_t$ over possible actions $A_t$ is then computed, reflecting the probability of selecting each action at step $t$ given the current states. Hidden representation $\mathbf{z}_t$ and policy distribution $\mathbf{d}_t$ are defined:

\begin{equation}
\label{eq:policy}
\begin{aligned}
&\hspace*{-0.1cm}
\mathbf{z}_t = \mathbf{W}_2 \text{ReLU}(\mathbf{W}_1 [\mathbf{s}_t^{k}||\mathbf{s}_t^{c}]),
\\
&\hspace*{-0.1cm}
\mathbf{d}_t = \pi_{\theta}(\cdot|s^k_t, s^c_t)=\text{softmax}(\mathbf{A}_t \mathbf{z}_t),
\end{aligned}
\end{equation}
where $\mathbf{W}_1, \mathbf{W}_2\in \mathbb{R}^{4d\times 4d}$ are the learnable weights, $\mathbf{A}_t\in \mathbb{R}^{|A_t|\times 4d}$ represent the embeddings of next possible actions $A_t$. The action $a_t$ at step $t$ is then selected as:

\begin{equation}
\label{eq:cluster_selection}
a_t\sim \text{categorical}(\textbf{d}_t).
\end{equation}

\noindent\textbf{Concept-Level Retrieval.}
Once the next semantic cluster $k_t$ is selected, the retriever proceeds to \underline{identify concepts within this cluster} \underline{to extend the reasoning paths.} This step grounds the high-level cluster selection in a concrete concept-to-concept transition within the medical KG. We mainly have two actions for retrieving concepts in the selected cluster $k_t$, as shown in Figure~\ref{fig:framework}(b).

\noindent\textit{\texttt{\underline{\textbf{Connect}}}.} 
If the selected cluster $k_t$ differs from $k_{t-1}$, we first establish a connection between them. Let $C_{\text{cand}}$ denote the set of concepts in $k_t$ that appear in previously explored paths $\{P^q_{t-1}\}_{q\in S^Q}$. For each path $P^q_{t-1}=\{c_0(q), ... r_{t-1}, c_{t-1}\}$, we select a connection point $c_{\text{leap}} \in C_{\text{cand}}$ and link the new cluster through $c_{\text{leap}}$. The point is chosen based on the maximum cosine similarity with the path embedding: $c_{\text{leap}}=\arg\max_{c\in C_{cand}}(\text{sim}(\textbf{c}, \textbf{P}^q_{t-1}))$, where $\mathbf{c}$ is the embedding of concept $c$, and $\mathbf{P}^q_{t-1}$ is the average embedding of concepts in path $P^q_{t-1}$. The path is then updated as $P^q_t = P^q_{t-1} \cup \{\text{``cluster leap''}, c_{\text{leap}}\}$.

\noindent\textit{\texttt{\underline{\textbf{Retrieve}}}.} 
 After establishing the connection, we retrieve new concepts by selecting $c_{\text{leap}}$'s neighbors $N(c_{\text{leap}})$ in cluster $k_t$. These new concepts provide semantically novel yet coherent information that extends and supports the prior reasoning path, thereby guiding the LLM to perform reasoning leaps and draw deeper inferences.

Let $\mathbf{S}^Q$ and $\mathbf{P}^q_t$ denote the average embeddings of keywords in $S^Q$ and concepts in path $P^q_t$, respectively. The new concept $c_t$ to be added to $P^q_t$ is selected as:

\begin{equation}
\label{eq:concept_selection}
c_t = \arg\max_{c' \in N(c_{\text{leap}})} 
\big[(\mathbf{c}', \mathbf{S}^Q), (\mathbf{c}', \mathbf{P}^q_t)\big]_{\text{sim}},
\end{equation}
where $[\cdot, \cdot]_{\text{sim}}$ denotes the average of the cosine similarities between the two pairs of embeddings. $\mathbf{c}'$ is the embedding of candidate concept $c'$. Therefore, the path is updated as $P^q_{t} = P^q_t \cup \{r_t, c_t\}$, where $r_t$ denotes the relation connecting $c_{\text{leap}}$ and $c_t$ in KG $\mathcal{G}$.

\subsection{GRO: Group-Based Retriever Optimization}
\label{subsec:gro}

To ensure that the retrieved paths can enhance LLM generation, a reward is provided when the reasoning paths reach the predefined length. This delayed feedback (episodic reward) allows the model to evaluate the overall quality of complete paths in supporting long-form instruction generation.

\noindent\textbf{Mixture of Rewards.}
The evaluation of each reasoning path $P$ is based on two criteria:
\ding{182} it contains concepts that appear in the ground-truth discharge instruction, directly contributing to accurate LLM outputs; and
\ding{183} it includes semantically related concepts that can guide the LLM toward generating relevant content.

To ensure these objectives are reflected in the episodic rewards, we adopt the following design. First, inspired by recent formulations of verifiable rewards~\cite{lambert2024t, guo2025deepseek}, we introduce a binary reward that assigns a value of 1 if the path contains any ground-truth concepts. Second, we incorporate a soft reward based on the embedding similarity between the concepts explored in $P$ and the ground-truth concepts $\hat{C}$. Thus, the reward for reasoning path $P$ is formulated as:

\begin{equation}
\label{eq:original_reward}
R_P = \sum_{c\in P}^{}\mathbb{I}\{c\in \hat{C}\} + \lambda\cdot\text{sim}(\mathbf{P},\mathbf{\hat{C}}),
\end{equation}
where $\lambda$ is a weighting factor. 
$\mathbf{P}$ and $\mathbf{\hat{C}}$ denote the average embeddings of concepts in $P$ and ground-truth set $\hat{C}$, respectively. $\text{sim}(\cdot, \cdot)$ represents the cosine similarity. $\mathbb{I}\{\cdot\}$ is the indicator function, which returns 1 if $c$ belongs to $\hat{C}$, and 0 otherwise.

\noindent\textbf{Group-Based Optimization.}
After each episode, the policy network is updated based on the rewards. Inspired by GRPO~\cite{shao2024deepseekmath}, we adopt its idea of using multiple rollouts per input to estimate the group-normalized reward. Therefore, we propose the \emph{GRO} mechanism (\underline{G}roup-Based \underline{R}etriever \underline{O}ptimization) to further improve the quality of retrieved paths under sparse episodic rewards. This also stabilizes learning by better attributing credit across entire paths.

Specifically, we perform a fixed number $G$ of retrieval processes for each patient. Let $\mathcal{P}^{(i)}$ denote the path set retrieved in the $i^{th}$ process. After $G$ retrievals, we obtain a reward set $\mathbf{R}=\{\mathcal{R}^{(1)}, ..., \mathcal{R}^{(G)}\}$, where $\mathcal{R}^{(i)}=\sum_{P\in\mathcal{P}^{(i)}}^{}R_P$. The group-normalized reward for each retrieval process is then formulated as:

\begin{equation}
\label{eq:group_reward}
\tilde{\mathcal{R}}^{(i)}=\frac{\mathcal{R}^{(i)}-\mu^{R}}{\sigma^{R}+\epsilon},
\end{equation}
where $\mu^{R}$ and $\sigma^{R}$ denote the mean and standard deviation of $\mathbf{R}$, respectively, and $\epsilon$ is a small constant for numerical stability.

The optimization aims to maximize the expected cumulative return.
We revise the REINFORCE algorithm~\citep{williams1992simple} by using discounted cumulative returns based on normalized rewards:

\begin{equation}
\label{eq:group_objective}
{\small
J(\theta)=\mathop{\mathbb{E}}_{\{\mathcal{P}^{(i)}\}^G_{i=1}\sim \pi_{\theta}}\left [ \frac{1}{G}\sum_{i=1}^{G} \sum_{t=0}^{T-1}\gamma^{(T-t)}\cdot\tilde{\mathcal{R}}^{(i)}\right],
}
\end{equation}
where $T$ is maximum path length and $\gamma\in [0,1]$ is the discount factor. To encourage exploration, the entropy term~\cite{williams1991function} is added: $\beta\mathcal{H}\left(\pi_{\theta}(\cdot|s^{(i)}_t)\right)$, where the state is $s^{(i)}_t = (s^{k{(i)}}_t, s^{c{(i)}}_t)$. $\mathcal{H}$ denotes policy entropy. $\beta\ge 0$ controls the exploration strength and is decayed during training. Let $\tilde{\mathcal{R}}^{(i)}_{t}=\gamma^{(T-t)}\cdot\tilde{\mathcal{R}}^{(i)}$ and $\mathcal{H}^{(i)}_t$ short for $\mathcal{H}\left(\pi_{\theta}(\cdot|s^{(i)}_t)\right)$.
The policy network $\pi_{\theta}$ is updated via the gradient of the objective:

\begin{multline}
\nabla_{\theta} J(\theta)=\\[6pt]
\mathbb{E}_{\{\mathcal{P}^{(i)}\}_{i=1}^{G}\sim \pi_{\theta}}
\Bigg[
\frac{1}{G}\sum_{i=1}^{G}\sum_{t=0}^{T-1}
\Big(
\nabla_{\theta}\log \pi_{\theta}(a^{(i)}_t|s^{(i)}_t)
\tilde{\mathcal{R}}^{(i)}_{t} \\[2pt]
+\beta\nabla_{\theta}\mathcal{H}^{(i)}_t
\Big)
\Bigg]
\label{eq:group_loss}
\end{multline}

Finally, given a well-trained retriever with policy $\hat{\pi}_{\theta}$, KG $\mathcal{G}$,  a new patient's pre-admission information $Q'$, and extracted keywords $S^{Q'}$, the reasoning paths $\{P^q\}_{q \in S^{Q'}} \sim \hat{\pi}_{\theta}$ are retrieved from $\mathcal{G}$. The LLM $\mathcal{M}$ then generates the ideal discharge instruction $\hat{\mathcal{I}}$ using our \emph{ReinRAG} model as follows:

\begin{equation}
\label{eq:generate}
\begin{aligned}
&\hspace*{0cm}
\hat{\mathcal{I}}=\operatorname{ReinRAG}(Q';\mathcal{M},\hat{\pi}_{\theta}, \mathcal{G})
\\
&\hspace*{0.2cm}
=\arg\max_{\mathcal{I}}\mathbb{P}_{\mathcal{M}}\Big(\mathcal{I} \:\big|\: Q', \{P^q\}_{q \in S^{Q'}} \sim \hat{\pi}_{\theta}\Big).
\end{aligned}
\end{equation}
\section{Experiments}

\subsection{Experimental Setup}

\noindent\textbf{Dataset and Preprocessing.}
We conduct experiments on a subset of MIMIC-IV-note~\cite{johnson2023mimic, goldberger2000components}, which contains 331,794 de-identified discharge summaries from 145,915 patients at the Beth Israel Deaconess Medical Center. We select 4,000 summaries, where 3,000 for training and 1,000 for testing. From each summary, we extract pre-admission information, including allergies, chief complaint, and history of present illness (HPI), which serves as both the model input and the prompt content for the LLMs. For the medical KG, we adopt the UMLS~\cite{bodenreider2004unified, UMLS2024AB}, a large-scale resource developed by the National Library of Medicine and structured as a KG with concepts, semantic relations, and semantic clusters (semantic groups). Following~\cite{gao2025leveraging}, we focus on SNOMED CT (Systematized Nomenclature of Medicine–Clinical Terms) concepts and use 107 diagnostic-related relations. Table~\ref{tb:data} summarizes data statistics.

\begin{table}[h]
\renewcommand{\arraystretch}{0.9}
\centering
\caption{Statistics of the medical KG and selected discharge instructions. ``Std.'' denotes the Standard Deviation, and ``TTR''represents the Type-Token Ratio.}
\label{tb:data}
\begin{tabular}{l|c||l|c}
\toprule[1.3pt]
\multicolumn{2}{c||}{\textbf{Medical KG}} & \multicolumn{2}{c}{\textbf{Discharge Instructions}} \\
\midrule[0.5pt]
\#Concepts & 443K & Avg. \#Words & 106.9 \\
\#Relation & 107 & Std. \#Words & 59.99 \\
\#Clusters & 15 & Avg. TTR & 0.7\\
\bottomrule[1.3pt]
\end{tabular}
\end{table}

\noindent\textbf{Keyword Extraction and Concept Matching}.
We use QuickUMLS~\cite{soldaini2016quickumls} to extract keywords from patient information and map them to UMLS concepts (focus on SNOMED CT). The best-matched concept for each keyword is selected. Neo4j is utilized to retrieve reasoning paths from the UMLS KG.

\begin{table*}[ht]
\centering
\renewcommand{\arraystretch}{1.1}
\setlength{\tabcolsep}{2pt}
\caption{CE evaluations (\%) of different models. ``N-gram'' and ``Concept'' refer to the keywords and medical concepts identified in the generated discharge instructions, respectively. ``J'' denotes Jaccard similarity, and ``HL'' represents Hamming loss. The best results are highlight in \textbf{bold}. The performance difference between baseline and \hlmethod{\emph{ReinRAG}} is reported in the ``$\Delta$'' column.}
\resizebox{\textwidth}{!}{
\begin{tabular}{lcTcTcTcTcTcTcTcTcTcT}
\toprule[1.6pt]
 & \multicolumn{20}{c}{\textbf{CE Metrics}} \\ 
\textbf{Model} \bm{$\downarrow$} & \multicolumn{10}{c}{\textbf{N-gram}} & \multicolumn{10}{c}{\textbf{Concept}} \\ \cmidrule(rl){2-11}\cmidrule(rl){12-21}
\textbf{Metric} \bm{$\to$} & P($\uparrow$) & $\Delta$ & R($\uparrow$) & $\Delta$ & F1($\uparrow$) & $\Delta$ & J($\uparrow$) & $\Delta$ & HL($\downarrow$) & $\Delta$
& P($\uparrow$) & $\Delta$ & R($\uparrow$) & $\Delta$ & F1($\uparrow$) & $\Delta$ & J($\uparrow$) & $\Delta$ & HL($\downarrow$) & $\Delta$\\
\midrule[1pt]
\multicolumn{21}{c}{\textit{\textbf{Vanilla LLMs}}}\\
\midrule[1pt]
LLaMA-3.1-8B      & 97.20 & \textit{(-1.6)} & 23.66 & \textit{(-11.2)} & 36.82 & \textit{(-13.0)} & 5.77 & \textit{(+0.5)} & 76.34 & \textit{(+11.2)} & 98.00 & \textit{(-1.2)} & 28.50 & \textit{(-12.2)} & 42.80 & \textit{(-13.2)} & 7.04 & \textit{(+0.6)} & 71.50 & \textit{(+12.2)}\\
\rowcolor{orange!10} Qwen2.5-7B  & 98.70 & \textit{(-0.1)} & 29.24 & \textit{(-5.6)} & 43.79 & \textit{(-6.0)} & \textbf{6.04} & \textit{(+0.8)} & 70.76 & \textit{(+5.6)} & 99.20 & \textit{(0.0)} & 34.74 & \textit{(-6.0)} & 50.14 & \textit{(-5.9)} & \textbf{7.41} & \textit{(+1.0)} & 65.26 & \textit{(+6.0)}\\
Qwen-UMLS-7B      & 86.40 & \textit{(-12.4)} & 14.14 & \textit{(-20.7)} & 23.01 & \textit{(-26.8)} & 4.00 & \textit{(-1.3)} & 85.86 & \textit{(+20.7)} & 91.20 & \textit{(-8.0)} & 18.20 & \textit{(-22.5)} & 28.69 & \textit{(-27.3)} & 5.23 & \textit{(-1.2)} & 81.80 & \textit{(+22.5)}\\
\rowcolor{orange!10} Mistral-7B-v0.3   & \textbf{99.00} & \textit{(+0.2)} & 28.61 & \textit{(-6.2)} & 42.94 & \textit{(-6.9)} & 5.71 & \textit{(+0.5)} & 71.39 & \textit{(+6.2)} & \textbf{99.60} & \textit{(+0.4)} & 34.24 & \textit{(-6.5)} & 49.56 & \textit{(-6.4)} & 7.04 & \textit{(+0.6)} & 65.76 & \textit{(+6.5)}\\
\midrule[1pt]
\multicolumn{21}{c}{\textit{\textbf{Medical-Domain LLMs}}}\\
\midrule[1pt]
ChatDoctor-7B  & 72.30 & \textit{(-26.5)} &  9.17 & \textit{(-25.6)} & 15.59 & \textit{(-34.2)} & 3.91 & \textit{(-1.3)} & 90.82 & \textit{(+25.6)} & 76.00 & \textit{(-23.2)} & 11.32 & \textit{(-29.4)} & 18.86 & \textit{(-37.1)} & 4.91 & \textit{(-1.5)} & 88.67 & \textit{(+29.4)}\\
Med-Alpaca-7B  & 82.50 & \textit{(-16.3)} & 13.30 & \textit{(-21.5)} & 21.85 & \textit{(-28.0)} & 4.31 & \textit{(-1.0)} & 86.69 & \textit{(+21.5)} & 85.80 & \textit{(-13.4)} & 16.27 & \textit{(-24.5)} & 26.09 & \textit{(-29.9)} & 5.43 & \textit{(-1.0)} & 83.72 & \textit{(+24.4)}\\
Meditron-7B  & 73.30 & \textit{(-25.5)} &  7.45 & \textit{(-27.4)} & 13.05 & \textit{(-36.8)} & 1.21 & \textit{(-4.0)} & 92.54 & \textit{(+27.4)} & 91.50 & \textit{(-7.7)} & 14.87 & \textit{(-25.9)} & 24.58 & \textit{(-31.4)} & 2.45 & \textit{(-4.0)} & 85.12 & \textit{(+25.9)}\\
Biomistral-7B  & 44.30 & \textit{(-54.5)} &  3.82 & \textit{(-31.0)} &  6.65 & \textit{(-43.2)} & 1.89 & \textit{(-3.4)} & 96.17 & \textit{(+31.0)} & 53.10 & \textit{(-46.1)} &  5.30 & \textit{(-35.4)} &  9.09 & \textit{(-46.9)} & 2.66 & \textit{(-3.8)} & 94.69 & \textit{(+35.4)}\\
PMC-LLaMA-13B  & 22.80 & \textit{(-76.0)} &  2.36 & \textit{(-32.5)} &  4.07 & \textit{(-45.8)} & 1.03 & \textit{(-4.2)} & 97.63 & \textit{(+32.4)} & 26.60 & \textit{(-72.6)} &  3.20 & \textit{(-37.5)} &  5.37 & \textit{(-50.6)} & 1.42 & \textit{(-5.0)} & 96.79 & \textit{(+37.5)}\\
MMed-Llama-3-8B  & 51.00 & \textit{(-47.8)} &  5.97 & \textit{(-28.8)} & 10.20 & \textit{(-39.6)} & 0.93 & \textit{(-4.3)} & 94.03 & \textit{(+28.8)} & 72.90 & \textit{(-26.3)} & 10.97 & \textit{(-29.8)} & 17.98 & \textit{(-38.0)} & 1.93 & \textit{(-4.5)} & 89.03 & \textit{(+29.8)}\\
\midrule[1pt]
\multicolumn{21}{c}{\textit{\textbf{Retrieval-Based Methods}}}\\
\midrule[1pt]
Random1hop & \\
\:\:+ LLaMA-3.1-8B & 98.10 & \textit{(-0.7)} & 26.72 & \textit{(-8.1)} & 40.60 & \textit{(-9.2)} & 5.79 & \textit{(+0.5)} & 73.28 & \textit{(+8.1)} & 98.40 & \textit{(-0.8)} & 31.82 & \textit{(-8.9)} & 46.63 & \textit{(-9.4)} & 7.05 & \textit{(+0.6)} & 68.18 & \textit{(+8.9)}\\
\:\:+ Qwen2.5-7B   & 98.70 & \textit{(-0.1)} & 28.97 & \textit{(-5.8)} & 43.52 & \textit{(-6.3)} & 5.76 & \textit{(+0.5)} & 71.03 & \textit{(+5.8)} & 98.90 & \textit{(-0.3)} & 34.32 & \textit{(-6.4)} & 49.73 & \textit{(-6.3)} & 7.04 & \textit{(+0.6)} & 65.68 & \textit{(+6.4)}\\
\:\:+ Qwen-UMLS-7B & 79.70 & \textit{(-19.1)} & 11.70 & \textit{(-23.1)} & 19.45 & \textit{(-30.4)} & 2.96 & \textit{(-2.3)} & 88.30 & \textit{(+23.1)} & 86.00 & \textit{(-13.2)} & 15.99 & \textit{(-24.7)} & 25.60 & \textit{(-30.4)} & 4.13 & \textit{(-2.3)} & 84.01 & \textit{(+24.7)}\\
\:\:+ Mistral-7B-v0.3 & 98.60 & \textit{(-0.2)} & 27.51 & \textit{(-7.3)} & 41.73 & \textit{(-8.1)} & 5.53 & \textit{(+0.3)} & 72.49 & \textit{(+7.3)} & 99.10 & \textit{(-0.1)} & 32.56 & \textit{(-8.2)} & 47.72 & \textit{(-8.3)} & 6.76 & \textit{(+0.3)} & 67.44 & \textit{(+8.2)}\\
\midrule[0.5pt]
Sim1hop & \\
\:\:+ LLaMA-3.1-8B  & 94.30 & \textit{(-4.5)} & 24.35 & \textit{(-10.5)} & 37.39 & \textit{(-12.4)} & 5.45 & \textit{(+0.2)} & 75.65 & \textit{(+10.5)} & 98.60 & \textit{(-0.6)} & 30.36 & \textit{(-10.4)} & 45.03 & \textit{(-11.0)} & 6.89 & \textit{(+0.5)} & 69.64 & \textit{(+10.4)}\\
\:\:+ Qwen2.5-7B  & \textbf{99.00} & \textit{(+0.2)} & 29.23 & \textit{(-5.6)} & 43.76 & \textit{(-6.1)} & 5.81 & \textit{(+0.5)} & 70.77 & \textit{(+5.6)} & 99.30 & \textit{(+0.1)} & 34.63 & \textit{(-6.1)} & 50.01 & \textit{(-6.0)} & 7.13 & \textit{(+0.7)} & 65.37 & \textit{(+6.1)}\\
\:\:+ Qwen-UMLS-7B  & 80.00 & \textit{(-18.8)} & 11.78 & \textit{(-23.0)} & 19.52 & \textit{(-30.3)} & 3.00 & \textit{(-2.3)} & 88.22 & \textit{(+23.0)} & 87.60 & \textit{(-11.6)} & 16.38 & \textit{(-24.3)} & 26.20 & \textit{(-29.8)} & 4.26 & \textit{(-2.2)} & 83.62 & \textit{(+24.4)}\\
\:\:+ Mistral-7B-v0.3 & 98.60 & \textit{(-0.2)} & 27.82 & \textit{(-7.0)} & 42.03 & \textit{(-7.8)} & 5.39 & \textit{(+0.1)} & 72.18 & \textit{(+7.0)} & 99.30 & \textit{(+0.1)} & 33.31 & \textit{(-7.4)} & 48.54 & \textit{(-7.5)} & 6.66 & \textit{(+0.2)} & 66.69 & \textit{(+7.4)}\\
\midrule[0.5pt]
DR.KNOWS&\\
\:\:+ Flan-T5-Large  & 32.80 & \textit{(-66.0)} &  2.97 & \textit{(-31.8)} &  5.20 & \textit{(-44.6)} & 1.41 & \textit{(-3.8)} & 97.03 & \textit{(+31.8)} & 54.00 & \textit{(-45.2)} &  5.13 & \textit{(-35.6)} &  8.88 & \textit{(-47.1)} & 2.60 & \textit{(-3.8)} & 94.87 & \textit{(+35.6)}\\
\:\:+ LLaMA-3.1-8B  & 93.20 & \textit{(-5.6)} & 15.65 & \textit{(-19.2)} & 25.84 & \textit{(-24.0)} & 2.27 & \textit{(-3.0)} & 84.35 & \textit{(+19.2)} & 98.10 & \textit{(-1.1)} & 23.44 & \textit{(-17.3)} & 36.55 & \textit{(-19.5)} & 3.44 & \textit{(-3.0)} & 76.56 & \textit{(+17.2)}\\
\:\:+ Mistral-7B-v0.3 & 91.40 & \textit{(-7.4)} & 13.59 & \textit{(-21.2)} & 22.87 & \textit{(-26.9)} & 3.71 & \textit{(-1.5)} & 86.41 & \textit{(+21.2)} & 94.50 & \textit{(-4.7)} & 17.55 & \textit{(-23.2)} & 28.61 & \textit{(-27.4)} & 4.91 & \textit{(-1.5)} & 82.45 & \textit{(+23.2)}\\
\midrule[1pt]
\multicolumn{21}{c}{\textit{\textbf{Our Model}}}\\
\midrule[1pt]
\rowcolor{blue!15} \emph{ReinRAG} (\textbf{ours}) & & & & & & & & & & & & & & & & & & & & \\
\rowcolor{blue!15}\:\:+ Mistral-7B-v0.3 & 98.80 & -- & \textbf{34.81} & -- & \textbf{49.82} & -- & 5.26 & -- & \textbf{65.19} & -- & 99.20 & -- & \textbf{40.73} & -- & \textbf{56.01} & -- & 6.42 & -- & \textbf{59.27} &--\\ 
\bottomrule[1.6pt]
\end{tabular}}
\label{tb:ce}
\end{table*}

\begin{table*}[ht]
\centering
\renewcommand{\arraystretch}{1.1}
\setlength{\tabcolsep}{0.01pt}
\caption{NLG evaluations (\%) of different models. ``RG'' and ``BL'' denote ROUGE and BLEU, respectively. ``MTR'' represents METEOR, and ``SBERT'' is short for Sentence-BERT. The best results are highlight in \textbf{bold}. The performance difference between baseline and \hlmethod{\emph{ReinRAG}} is reported in the ``$\Delta$'' column.}
\resizebox{\textwidth}{!}{
\begin{tabular}{lcTcTcTcTcTcTcTcTcTcTcT}
\toprule[1.6pt]
\textbf{Model} \bm{$\downarrow$} & \multicolumn{20}{c}{\textbf{NLG Metrics}} \\
\textbf{Metric} \bm{$\to$} & RG-1($\uparrow$) & $\Delta$ & RG-2($\uparrow$) & $\Delta$ & RG-L($\uparrow$) & $\Delta$ & BL-1($\uparrow$) & $\Delta$ & BL-2($\uparrow$) & $\Delta$ & P$_\text{BERT}$($\uparrow$) & $\Delta$ & R$_\text{BERT}$($\uparrow$) & $\Delta$ & F1$_\text{BERT}$($\uparrow$) & $\Delta$ & MTR($\uparrow$) & $\Delta$ & SBERT($\uparrow$) & $\Delta$ \\ 
\midrule[1pt]
\multicolumn{21}{c}{\textit{\textbf{Vanilla LLMs}}} \\ 
\midrule[1pt]
LLaMA-3.1-8B      & 21.28 & \textit{(-0.3)} & 3.14 & \textit{(-1.1)} & 11.04 & \textit{(-1.0)} & 14.89 & \textit{(+4.0)} & 6.12 & \textit{(+0.5)} & 80.85 & \textit{(-0.1)} & 81.82 & \textit{(-1.7)} & 81.32 & \textit{(-0.9)} & 22.75 & \textit{(-1.3)} & 46.18 & \textit{(-9.1)} \\
Qwen2.5-7B        & 20.30 & \textit{(-1.3)} & 3.75 & \textit{(-0.5)} & 10.81 & \textit{(-1.3)} & 13.32 & \textit{(+2.5)} & 6.13 & \textit{(+0.5)} & 80.24 & \textit{(-0.7)} & 82.50 & \textit{(-1.1)} & 81.34 & \textit{(-0.9)} & 24.02 & \textit{(-0.1)} & 47.74 & \textit{(-7.5)} \\
Qwen-UMLS-7B      & 14.96 & \textit{(-6.6)} & 1.82 & \textit{(-2.5)} &  8.36 & \textit{(-3.7)} & 10.14 & \textit{(-0.7)} & 4.08 & \textit{(-1.5)} & 78.58 & \textit{(-2.4)} & 80.80 & \textit{(-2.8)} & 79.63 & \textit{(-2.6)} & 16.16 & \textit{(-7.9)} & 39.27 & \textit{(-16.0)} \\
Mistral-7B-v0.3   & 20.11 & \textit{(-1.5)} & 3.26 & \textit{(-1.0)} & 10.33 & \textit{(-1.7)} & 12.94 & \textit{(+2.1)} & 5.38 & \textit{(-0.2)} & 80.17 & \textit{(-0.8)} & 82.23 & \textit{(-1.3)} & 81.18 & \textit{(-1.0)} & 23.30 & \textit{(-0.8)} & 43.83 & \textit{(-11.4)} \\
\midrule[1pt]
\multicolumn{21}{c}{\textit{\textbf{Medical-Domain LLMs}}} \\
\midrule[1pt]
\rowcolor{orange!10} ChatDoctor-7B   & 16.46 & \textit{(-5.1)} & 1.85 & \textit{(-2.4)} &  9.53 & \textit{(-2.5)} & \textbf{19.72} & \textit{(+8.9)} & \textbf{6.97} & \textit{(+1.4)} & \textbf{81.82} & \textit{(+0.9)} & 80.58 & \textit{(-3.0)} & 81.17 & \textit{(-1.0)} & 13.49 & \textit{(-10.6)} & 30.93 & \textit{(-24.3)} \\
Med-Alpaca-7B     & 16.99 & \textit{(-4.6)} & 1.85 & \textit{(-2.4)} &  9.64 & \textit{(-2.4)} & 16.78 & \textit{(+5.9)} & 5.64 & \textit{(+0.1)} & 81.08 & \textit{(+0.1)} & 80.11 & \textit{(-3.5)} & 80.56 & \textit{(-1.7)} & 15.75 & \textit{(-8.3)} & 36.00 & \textit{(-19.2)} \\
Meditron-7B       &  9.94 & \textit{(-11.6)} & 0.76 & \textit{(-3.5)} &  5.66 & \textit{(-6.4)} &  6.61 & \textit{(-4.3)} & 2.09 & \textit{(-3.5)} & 75.72 & \textit{(-5.3)} & 79.48 & \textit{(-4.1)} & 77.54 & \textit{(-4.7)} & 15.21 & \textit{(-8.9)} & 15.96 & \textit{(-39.3)} \\
Biomistral-7B     & 10.40 & \textit{(-11.2)} & 0.76 & \textit{(-3.5)} &  6.80 & \textit{(-5.3)} &  7.99 & \textit{(-2.9)} & 2.25 & \textit{(-3.3)} & 79.70 & \textit{(-1.3)} & 76.72 & \textit{(-6.8)} & 78.15 & \textit{(-4.1)} &  7.36 & \textit{(-16.7)} & 20.81 & \textit{(-34.4)} \\
PMC-LLaMA-13B     &  5.35 & \textit{(-16.2)} & 0.47 & \textit{(-3.8)} &  3.54 & \textit{(-8.5)} &  2.80 & \textit{(-8.1)} & 0.90 & \textit{(-4.7)} & 68.14 & \textit{(-12.8)} & 66.37 & \textit{(-17.2)} & 67.22 & \textit{(-15.0)} &  3.91 & \textit{(-20.2)} & 13.24 & \textit{(-42.0)} \\
MMed-Llama-3-8B   &  6.06 & \textit{(-15.5)} & 0.44 & \textit{(-3.8)} &  3.51 & \textit{(-8.6)} &  4.70 & \textit{(-6.2)} & 1.41 & \textit{(-4.2)} & 71.17 & \textit{(-9.8)} & 77.19 & \textit{(-6.4)} & 74.01 & \textit{(-8.2)} & 10.14 & \textit{(-13.9)} & 15.25 & \textit{(-40.0)} \\
\midrule[1pt]
\multicolumn{21}{c}{\textit{\textbf{Retrieval-Based Methods}}} \\
\midrule[1pt]
Random1hop & & & & & & & & & & & & & & & & & & & & \\ 
\:\:+ LLaMA-3.1-8B    & 20.08 & \textit{(-1.5)} & 3.24 & \textit{(-1.0)} & 10.68 & \textit{(-1.4)} & 13.17 & \textit{(+2.3)} & 5.64 & \textit{(+0.1)} & 80.80 & \textit{(-0.2)} & 82.22 & \textit{(-1.3)} & 81.49 & \textit{(-0.7)} & 22.59 & \textit{(-1.5)} & 48.51 & \textit{(-6.7)} \\
\:\:+ Qwen2.5-7B      & 19.71 & \textit{(-1.9)} & 3.54 & \textit{(-0.7)} & 10.50 & \textit{(-1.6)} & 12.53 & \textit{(+1.7)} & 5.63 & \textit{(0.0)} & 80.28 & \textit{(-0.7)} & 82.41 & \textit{(-1.2)} & 81.31 & \textit{(-0.9)} & 23.50 & \textit{(-0.6)} & 48.38 & \textit{(-6.9)} \\
\:\:+ Qwen-UMLS-7B    & 13.50 & \textit{(-8.1)} & 1.51 & \textit{(-2.8)} &  7.72 & \textit{(-4.3)} &  8.65 & \textit{(-2.2)} & 3.32 & \textit{(-2.3)} & 77.97 & \textit{(-3.0)} & 80.03 & \textit{(-3.5)} & 78.93 & \textit{(-3.3)} & 14.64 & \textit{(-9.4)} & 35.75 & \textit{(-19.5)} \\
\:\:+ Mistral-7B-v0.3 & 19.70 & \textit{(-1.9)} & 3.08 & \textit{(-1.2)} & 10.21 & \textit{(-1.9)} & 12.49 & \textit{(+1.6)} & 5.12 & \textit{(-0.5)} & 80.19 & \textit{(-0.8)} & 81.89 & \textit{(-1.7)} & 81.02 & \textit{(-1.2)} & 22.86 & \textit{(-1.2)} & 43.83 & \textit{(-11.4)} \\
\midrule[0.5pt]
Sim1hop & & & & & & & & & & & & & & & & & & & & \\ 
\:\:+ LLaMA-3.1-8B    & 20.32 & \textit{(-1.3)} & 3.19 & \textit{(-1.1)} & 10.71 & \textit{(-1.4)} & 13.41 & \textit{(+2.5)} & 5.62 & \textit{(0.0)} & 80.76 & \textit{(-0.2)} & 82.30 & \textit{(-1.3)} & 81.50 & \textit{(-0.7)} & 22.50 & \textit{(-1.6)} & 47.89 & \textit{(-7.4)} \\
\:\:+ Qwen2.5-7B      & 19.72 & \textit{(-1.9)} & 3.55 & \textit{(-0.7)} & 10.52 & \textit{(-1.6)} & 12.57 & \textit{(+1.7)} & 5.67 & \textit{(+0.1)} & 80.26 & \textit{(-0.7)} & 82.38 & \textit{(-1.2)} & 81.29 & \textit{(-0.9)} & 23.57 & \textit{(-0.5)} & 48.31 & \textit{(-7.0)} \\
\:\:+ Qwen-UMLS-7B    & 13.65 & \textit{(-7.9)} & 1.56 & \textit{(-2.7)} &  7.81 & \textit{(-4.3)} &  8.53 & \textit{(-2.3)} & 3.29 & \textit{(-2.3)} & 77.87 & \textit{(-3.1)} & 80.04 & \textit{(-3.5)} & 78.89 & \textit{(-3.3)} & 14.76 & \textit{(-9.3)} & 36.21 & \textit{(-19.0)} \\
\:\:+ Mistral-7B-v0.3 & 19.22 & \textit{(-2.4)} & 3.05 & \textit{(-1.2)} & 10.04 & \textit{(-2.0)} & 12.09 & \textit{(+1.2)} & 4.99 & \textit{(-0.6)} & 80.09 & \textit{(-0.9)} & 81.88 & \textit{(-1.7)} & 80.96 & \textit{(-1.3)} & 22.80 & \textit{(-1.3)} & 43.85 & \textit{(-11.4)} \\
\midrule[0.5pt]
DR.KNOWS & & & & & & & & & & & & & & & & & & & & \\ 
\:\:+ Flan-T5-Large   &  6.75 & \textit{(-14.8)} & 0.49 & \textit{(-3.8)} &  4.65 & \textit{(-7.4)} &  9.04 & \textit{(-1.8)} & 3.26 & \textit{(-2.3)} & 76.53 & \textit{(-4.4)} & 78.42 & \textit{(-5.1)} & 77.43 & \textit{(-4.8)} &  5.78 & \textit{(-18.3)} & 23.52 & \textit{(-31.7)} \\
\:\:+ LLaMA-3.1-8B    &  8.44 & \textit{(-13.1)} & 0.83 & \textit{(-3.5)} &  4.78 & \textit{(-7.3)} &  5.36 & \textit{(-5.5)} & 1.85 & \textit{(-3.7)} & 76.42 & \textit{(-4.5)} & 80.98 & \textit{(-2.6)} & 78.61 & \textit{(-3.6)} & 14.42 & \textit{(-9.7)} & 38.54 & \textit{(-16.7)} \\
\:\:+ Mistral-7B-v0.3 & 15.57 & \textit{(-6.0)} & 1.59 & \textit{(-2.7)} &  8.93 & \textit{(-3.1)} & 13.13 & \textit{(+2.3)} & 4.36 & \textit{(-1.2)} & 79.65 & \textit{(-1.3)} & 80.74 & \textit{(-2.8)} & 80.74 & \textit{(-1.5)} & 17.21 & \textit{(-6.9)} & 43.71 & \textit{(-11.5)} \\
\midrule[1pt]
\multicolumn{21}{c}{\textit{\textbf{Our Model}}} \\
\midrule[1pt]
\rowcolor{blue!15} \emph{ReinRAG} (\textbf{ours}) & & & & & & & & & & & & & & & & & & & & \\
\rowcolor{blue!15} \:\:+ Mistral-7B-v0.3 & \textbf{21.57} & -- & \textbf{4.28} & -- & \textbf{12.07} & -- & 10.87 & -- & 5.58 & -- & 80.97 & -- & \textbf{83.56} & -- & \textbf{82.22} & -- & \textbf{24.07} & -- & \textbf{55.24} & --\\
\bottomrule[1.6pt]
\end{tabular}}
\label{tb:nlg}
\end{table*}

\noindent\textbf{Baselines.}
We compare with following baselines:

\begin{itemize}[leftmargin=*]
    \item \textbf{Vanilla LLMs} include LLaMA-3.1-8B-Instruct~\cite{dubey2024llama}, Qwen2.5-7B-Instruct~\cite{yang2024qwen2}, Qwen-UMLS-7B-Instruct~\cite{QwenUMLS7BInstruct}, Mistral-7B-Instruct-v0.3~\cite{jiang2023mistral}, using pre-admission data as prompt for generation.
    \item \textbf{Medical-Domain LLMs} include LLMs pre-trained or instruction-tuned on biomedical literature, clinical notes, or medical QA corpora, including ChatDoctor-7B~\cite{li2023chatdoctor}, Med-Alpaca-7B~\cite{shu2023visual}, Meditron-7B~\cite{chen2023meditron}, Biomistral-7B~\cite{DBLP:conf/acl/LabrakBMGRD24}, PMC-LLaMA-13B~\cite{wu2024pmc}, and MMed-Llama-3-8B~\cite{qiu2024towards}.
    \item \textbf{Retrieved-Based Methods} consider two one-hop neighbor retrieval baselines. Both identify KG concepts structurally connected to keywords extracted from the pre-admission information: one randomly selects one-hop neighbors, denoted as ``\textbf{Random1hop}'', and the other selects those most semantically similar to the full pre-admission input, denoted as ``\textbf{Sim1hop}''. Both baselines retrieve from the KG without performing reasoning leaps or structuring the retrieved information into paths. We also compare with \textbf{DR.KNOWS}~\cite{gao2025leveraging}, which performs path-based retrieval on the KG. The retrieved concepts and original input are used to prompt LLMs for generation.
\end{itemize}

\noindent\textbf{Evaluation Metrics.}
Models are evaluated with two types of metrics to compare generated and ground-truth discharge instructions:

\begin{itemize}[leftmargin=*]
    \item \textbf{Clinical Efficacy (CE)}: We assess the correctness of the generated instructions by matching keyword (\underline{N-gram level}) and SNOMED CT concepts (\underline{concept level}) with concepts from ground-truth instructions, using precision, recall, F1 score, Hamming loss, and Jaccard similarity. These metrics evaluate the correctness of medically relevant word generation.

    \item \textbf{Natural Language Generation (NLG)}: We report ROUGE-1/2/L~\cite{lin2004rouge}, BLEU-1/2~\cite{papineni2002bleu}, METEOR~\cite{denkowski2011meteor}, BERTScore (F1), and Sentence-BERT~\cite{reimers2019sentence} similarity scores to measure the fluency and semantic consistency of the generation.
\end{itemize}

\begin{figure*}[h]
\centering
\graphicspath{{figs/}}
\centering
\subfloat
  [Performance with different number of retrieval process ($G$) per patient.]{\label{subfig:extra_analysis_group_size}\includegraphics[width=0.9\textwidth]{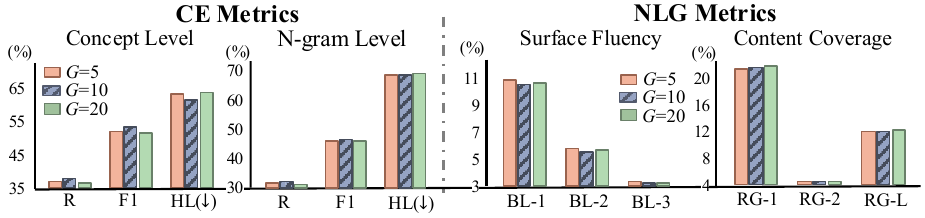}}
\\
\subfloat
  [Performance with different number of paths ($N_P$) prompted to the LLM.]{\label{subfig:extra_analysis_path_num}\includegraphics[width=0.9\textwidth]{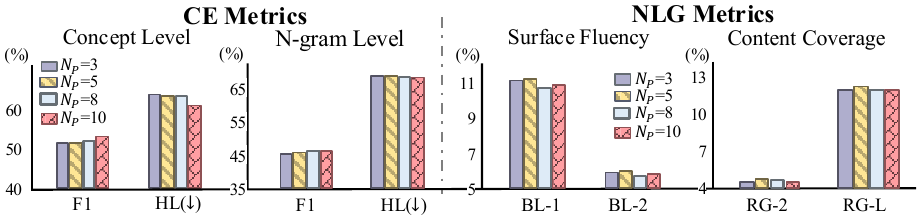}}
\caption{Parameter sensitivity analysis of \emph{ReinRAG} with Mistral-7B-Instruct-v0.3.}
\label{fig:extra_analysis}
\end{figure*}

\begin{figure}[h]
\centering
\graphicspath{{figs/}}
\centering
\subfloat
  [CE Evaluation.]{\label{subfig:ablation_ce}\includegraphics[width=0.25\textwidth]{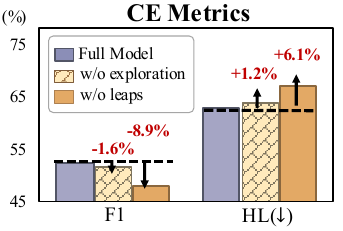}}
\subfloat
  [NLG Evaluation.]{\label{subfig:ablation_nlg}\includegraphics[width=0.25\textwidth]{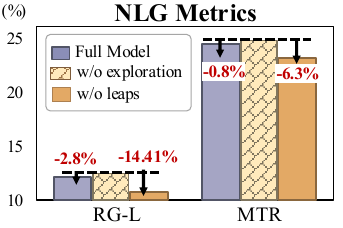}}
\caption{Ablation study of \emph{ReinRAG} with Mistral.}
\label{fig:extra_analysis_ablation}
\end{figure}

\subsection{Comparison Performance}
Table~\ref{tb:ce} and Table~\ref{tb:nlg} report the CE and NLG performance of all models, respectively. Key findings are summarized below:

\noindent\textbf{\underline{Clinical Accuracy and Noise Sensitivity.}} In Table~\ref{tb:ce}, \emph{ReinRAG} achieves comparable precision, while outperforming vanilla LLaMA and the best baseline (Qwen2.5) by at least 12\% and 6\%, respectively, in both recall and F1 score. Interestingly, vanilla LLMs sometimes outperform retrieval-based baselines. This suggests that simply retrieving information directly related to pre-admission data can sometimes degrade LLM performance. Medical-domain LLMs, which are pre-trained on clinical corpora for short-form tasks, also fail to improve performance. In contrast, \emph{ReinRAG} achieves the highest F1 score while also reducing Hamming loss by at least 12\% and 5\% compared to vanilla LLaMA and the best baselines (Qwen2.5), respectively. This indicates that incorporating RL into retrieval can effectively guide LLMs toward accurate long-form generation rather than hindering it.

\noindent\textbf{\underline{Semantic Consistency.}} In Table~\ref{tb:nlg}, \emph{ReinRAG} achieves the highest scores on most metrics. This indicates that our generation preserves the core meaning of ground-truth instructions with less irrelevant descriptions. Although it obtains a lower BLEU-2 score, the highest ROUGE-L, BERTScore (F1$_\text{BERT}$), METEOR and Sentence-BERT similarity scores confirm that \emph{ReinRAG} produces outputs that remain semantically similar to ground truths at the paragraph level. This suggests that our generation better captures longer-range overlaps and adheres more closely to ground truths.

\noindent\textbf{\underline{Effectiveness of Reasoning Leaps.}} Similar to our method, DR.KNOWS~\cite{gao2025leveraging} also retrieves paths from the KG to prompt LLMs. However, its retrieval is limited to concepts directly connected to the prompt content. This restricts its ability to reason across distant semantic information. As a result, it underperforms \emph{ReinRAG} across all metrics. This demonstrates that \emph{ReinRAG}'s adaptive control of reasoning granularity, which allows reasoning leaps, can form more effective paths to better guide LLM generation.

\subsection{Parameter Sensitivity Analysis.}

To evaluate the impact of the number of retrieval processes ($G$ in Eq.~\ref{eq:group_objective}) and the number of explored paths prompted to the LLM (denoted as $N_P$), we vary these parameters to examine the performance. In Figure~\ref{subfig:extra_analysis_group_size}, setting $G$ to 10 achieves better CE performance. Increasing $G$ to 20 slightly improves ROUGE scores but decreases F1, suggesting that excessive retrievals may introduce noise and harm medical concept correctness for individual patients, despite slightly improve overall content coverage. Figure~\ref{subfig:extra_analysis_path_num} indicates that a larger number of prompted paths generally improve CE metrics, but too many paths may also reduce semantic consistency in LLM generation. These results highlight the importance of properly setting both the number of retrievals and prompted paths to balance CE and NLG performance.

\begin{figure}[t]
\graphicspath{{figs/}}
\begin{center}
\includegraphics[width=0.48\textwidth]{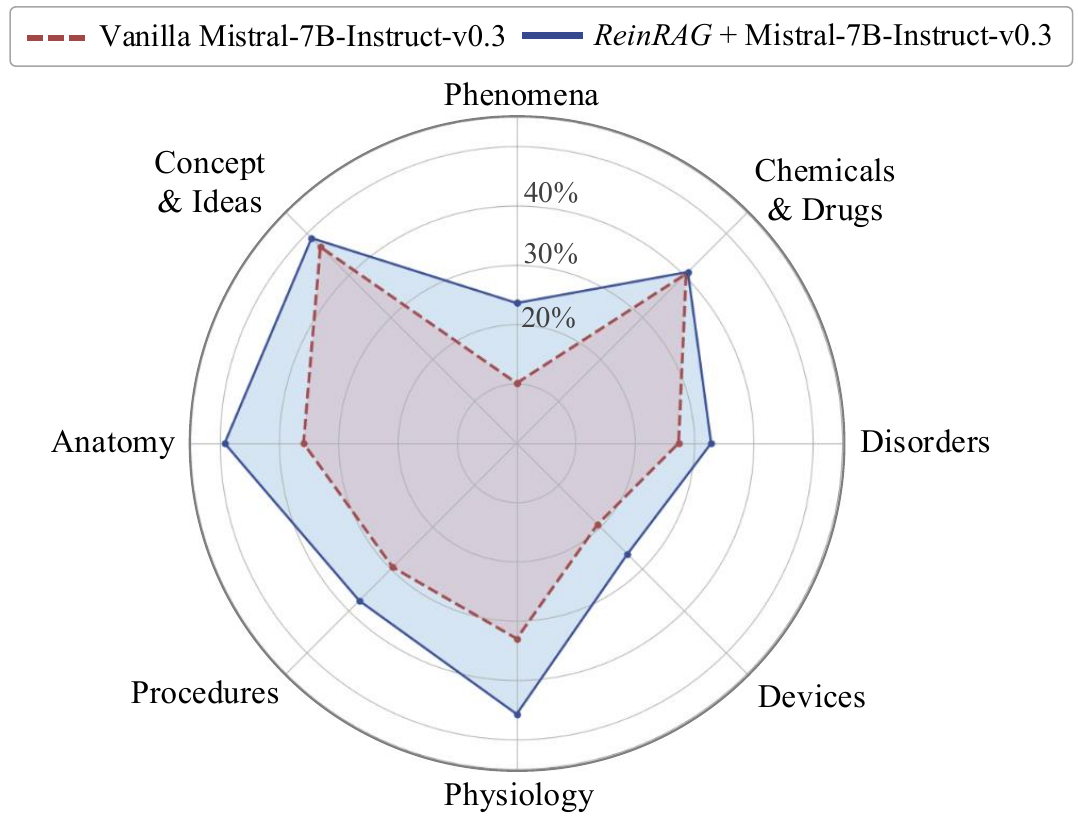}
\end{center}
\caption{Recall of vanilla Mistral-7B-Instruct-v0.3 and our \emph{ReinRAG} model across semantic clusters in the UMLS KG.}
\label{fig:cluster_performance}
\end{figure}

\subsection{Ablation Study}

To evaluate the design in our RL-based retriever, we conduct an ablation study by removing (i) the exploration ability of \emph{ReinRAG} (the entropy term in Eq.~\ref{eq:group_loss}) and (ii) reasoning leaps during retrieval, referred to ``w/o exploration'' and ``w/o leaps'', respectively. Results in Figure~\ref{fig:extra_analysis_ablation} indicate that removing reasoning leaps significantly degrades both CE and NLG performance. Removing exploration ability slightly improves ROUGE-L and METEOR but leads to lower F1 score and higher Hamming loss compared to the full \emph{ReinRAG}. These results suggest that allowing reasoning leaps effectively guides the LLM toward broader reasoning granularity, helping it generate more accurate information. Meanwhile, the exploration ability of \emph{ReinRAG}, despite slightly sacrificing the semantic consistency with ground truths, improves the LLM to generate more accurate concepts. Proper tuning the exploration strength can further balance and enhance the performance, demonstrating the effectiveness of the \emph{ReinRAG} design.

\subsection{Impact Across Semantic Clusters}

To analyze which aspects of generation benefit from \emph{ReinRAG}, we compare the recall of medical concepts generated by \emph{ReinRAG} and vanilla Mistral across eight representative semantic clusters. The results are shown in Figure~\ref{fig:cluster_performance}.

\noindent\underline{\textit{Limited Impact in Well-Covered Semantics.}} In clusters such as \textit{Concepts \& Ideas} and \textit{Chemicals \& Drugs}, \emph{ReinRAG} shows similar performance to vanilla Mistral. These clusters primarily include non-critical terms (e.g., ``Dosing instruction fragment'') or explicitly mentioned pre-admission medications. Thus, the vanilla Mistral already achieves high recall in these clusters, suggesting that \emph{ReinRAG} contributes less in these semantic information.

\noindent\underline{\textit{Improved Recall in Information-Sparse Clusters.}} \emph{ReinRAG} significantly improves recall in clusters like \textit{Anatomy}, \textit{Procedures}, \textit{Physiology}, and \textit{Phenomena}, which include concepts related to body parts, diagnoses, treatments, organ functions, and physiological phenomena. These types of information are typically gathered during a patient’s hospital stay and are often underrepresented or implicit in the pre-admission data. This demonstrate that \emph{ReinRAG} effectively bridges the information gap by retrieving reasoning paths from the KG based on known clues.

\begin{table}[t]
\renewcommand{\arraystretch}{0.4}
\caption{Medical professionals' feedback on discharge instructions generated by Vanilla-Mistral-7B-Instruct-v0.3 and our \emph{ReinRAG}.}
\label{tb:expert_feedback}
\centering
\small
\begin{tabularx}{0.5\textwidth}{c|L|L}
\toprule[0.8pt]
\textbf{Model} & \textbf{Strengths} & \textbf{Weakness} \\
\midrule[1pt]
\textbf{Vanilla}  & 
\textit{``The care suggestions are detailed and comprehensive, and the instructions are highly related to patients' pre-admission information.''}
&  
\textit{``Unrelated medications and diagnostic errors often occurs, such as inappropriate medication or diet suggestions. Most of the diagnostic logic is messy and irrelevant.''}
\\
\midrule[0.8pt]
\textbf{\emph{ReinRAG}} & 
\textit{``The instructions are more concise and logical, focusing on the core diagnosis and treatments. The number of wrong diagnoses is relatively low.''}
& 
\textit{``The instructions are sometimes unclear. There are occasional information errors and omissions in a few cases, though key concepts are mentioned.''}
\\
\bottomrule[1pt]
\end{tabularx}
\end{table}

\subsection{Human Evaluation}

To verify whether \emph{ReinRAG} can assist clinical practice, we invite two medical processionals to conduct a human evaluation. They review 20 instructions generated by Vanilla Mistral and our \emph{ReinRAG}.

In Table~\ref{tb:expert_feedback}, we present representative comments from two medical professionals after they review 20 patient cases. The feedback reveals that vanilla Mistral tends to provide more comprehensive discharge information but often generates irrelevant or event incorrect instructions. In contrast, while \emph{ReinRAG}'s generation occasionally lacks detailed descriptions, the outputs are more accurate and logically reasoned. This suggests that vanilla Mistral, without guidance of our reasoning paths, may provide abundant medical information but often in the wrong direction, failing to align with the patients' actual clinical needs.

\begin{figure}[t]
\graphicspath{{figs/}}
\begin{center}
\includegraphics[width=0.48\textwidth]{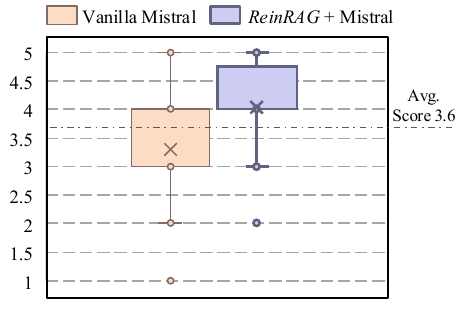}
\end{center}
\caption{Overall human evaluation performance of Vanilla Mistral-7B-Instruct-v0.3 and \emph{ReinRAG}. Scores range from 1 to 5, with higher scores indicating better performance.}
\label{fig:human_overall_score}
\end{figure}

\begin{figure}[t]
\graphicspath{{figs/}}
\begin{center}
\includegraphics[width=0.48\textwidth]{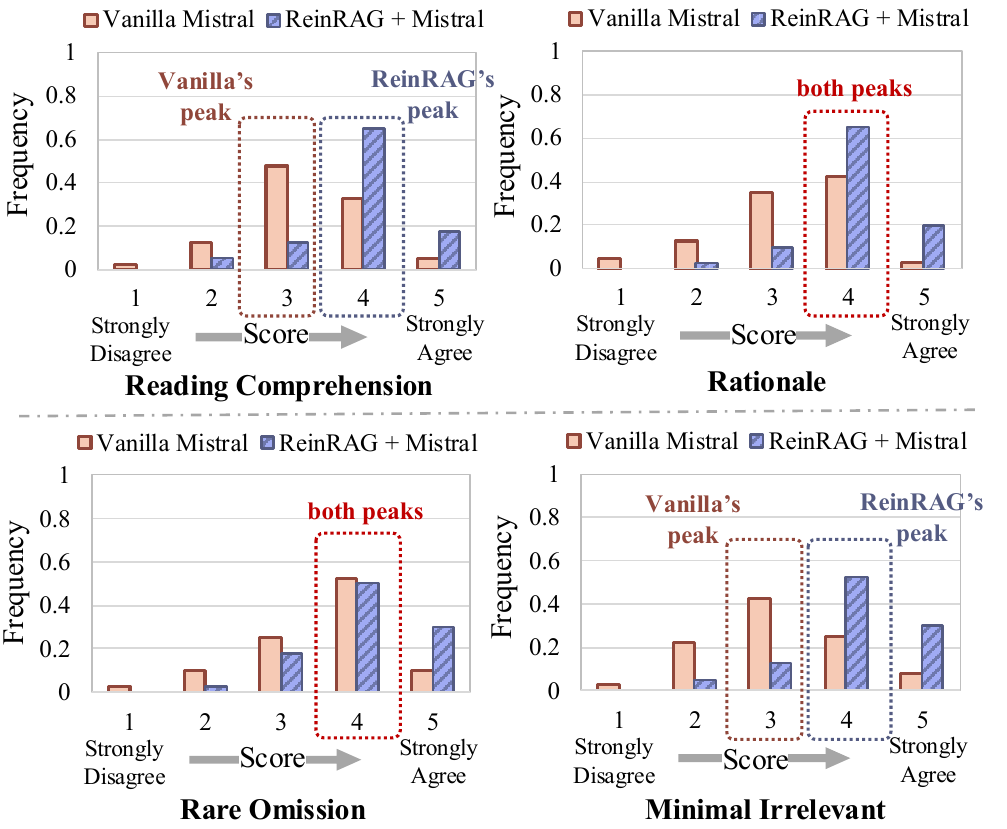}
\end{center}
\caption{Evaluation scores over four aspects.}
\label{fig:human_detail_score}
\end{figure}

Moreover, inspired by the evaluation designed in~\cite{gao2025leveraging}, two medical professionals assess the generation based on following aspects: (1) reading comprehension, (2) rationale instructions, (3) rare omission of critical information and (4) minimal irrelevant information. The scores range from 1 to 5, representing strongly disagree, disagree, neutral, agree, and strongly agree, respectively. The overall evaluation scores of both methods are shown in Figure~\ref{fig:human_overall_score}, and the detailed scores of different aspects are provided in Figure~\ref{fig:human_detail_score}.

As illustrated in Figure~\ref{fig:human_overall_score}, \emph{ReinRAG} outperforms the vanilla Mistral model, with not only higher average scores but also a narrower value range, indicating more consistent evaluations. Regarding the four evaluation aspects shown in Figure~\ref{fig:human_detail_score}, \emph{ReinRAG}'s score distributions are consistently skewed toward higher values.
Among the four aspects, the smallest gap between the two models is observed in the evaluation of rare omission of critical information. In contrast, for the rationale of instructions, \emph{ReinRAG} more frequently achieves higher scores. The differences in score distributions become more pronounced in the aspects of reading comprehension and minimal irrelevant information, suggesting that \emph{ReinRAG} generates outputs that are not only more faithful but also more concise and focused. 

This human evaluation highlights the potential of \emph{ReinRAG} to assist clinicians as a reference for early clinical decision-making.

\section{Conclusion}

This paper introduces \emph{ReinRAG}, a novel RL–based retrieval leveraging reasoning paths to guide LLMs in generating discharge instructions using only pre-admission data. By controlling the reasoning granularity through reasoning leaps and utilizing group-normalized rewards via the proposed GRO, \emph{ReinRAG} effectively retrieves high-quality reasoning paths. Experimental results on the MIMIC-IV-Note dataset show that \emph{ReinRAG} outperforms baseline approaches in both clinical efficacy and natural language generation.

\section*{Limitations}

While \emph{ReinRAG} shows strong performance, several limitations should be acknowledged. First, although our experiments demonstrate improvements in clinical concept coverage and generation quality, more comprehensive human evaluations by physicians are needed to strengthen performance evaluation. Second, the current fixed-length retrieval in \emph{ReinRAG} may limit adaptability to varying patient complexity. Incorporating adaptive reasoning lengths based on prompt context remains an important direction for future work.

\section*{Ethical Statement}

All datasets used in this research are publicly available and are obtained according to respective data usage policies. The data is de-identified, and we do not attempt to re-identify any individuals.

\bibliography{anthology}
\bibliographystyle{acl_natbib}

\appendix
\section{Implementation Details}
\label{appendix:implement_detail}

\subsection{Hyperparameter Settings}

For model training, the maximum number of retrieval steps is set to 5, and the embedding dimension is 768.
We train the model for 500 epochs with a batch size of 48. The discount factor ($\gamma$ in Eq.~\ref{eq:group_objective}) is set to 0.1, and the weight $\lambda$ (Eq.~\ref{eq:original_reward}) is set to 10. The number of retrieval processes per sample ($G$ in Eq.~\ref{eq:group_objective}) and the number of reasoning paths prompted to the LLM are both set to 10.

\subsection{Prompt of \emph{ReinRAG}}

\tcbset{
    promptstyle/.style={
        colback=gray!5,
        colframe=black!70,
        fonttitle=\bfseries,
        boxrule=0.5mm,
        sharp corners,
        breakable,
        width=1\linewidth
    }
}
\begin{tcolorbox}[promptstyle, title=\emph{ReinRAG} Prompt]
\small
\ttfamily
You are a doctor tasked with generating discharge instructions for patients. You are equipped with a medical knowledge graph. Always provide clear, actionable advice and explain medical terms for patient understanding.
\\

Below provides the [EXAMPLE PATIENT CONDITION], [EXAMPLE RETRIEVED REASONING PATHS] from the medical knowledge graph, and the corresponding [EXAMPLE DISCHARGE INSTRUCTIONS]. Please use this example as a guide to generate [NEW DISCHARGE INSTRUCTIONS] for the new patient based on the provided [NEW PATIENT CONDITION] and [NEW RETRIEVED REASONING PATHS] from the knowledge graph. 
\\

Note that the path format of both the [EXAMPLE RETRIEVED REASONING PATHS] and [NEW RETRIEVED REASONING PATHS] follows this structure: concept [semantic group] $\rightarrow$ relation $\rightarrow$ concept [semantic group] $\rightarrow$ ...
\\

Please write the [NEW DISCHARGE INSTRUCTIONS] in a single, flowing paragraph format without using separate titles or headings. Address the following aspects: medications, dietary recommendations, activity level adjustments, and any specific precautions related to the Allergies, Chief Complaint, and History of Present Illness, without the greeting sentences.
Ensure the [NEW DISCHARGE INSTRUCTIONS] are clearly structured, with actionable advice and all medical terms explained for the patient's understanding.
\\

\begin{verbatim}
[EXAMPLE PATIENT CONDITION]:
{example_patient_condition}

[EXAMPLE RETRIEVED REASONING PATHS]:
{example_retrieved_reasoning_paths}

[EXAMPLE DISCHARGE INSTRUCTIONS]:
{example_discharge_instructions}

[NEW PATIENT CONDITION]:
{new_patient_condition}

[NEW RETRIEVED REASONING PATHS]:
{new_retrieved_reasoning_paths}

[NEW DISCHARGE INSTRUCTIONS]:

\end{verbatim}
\end{tcolorbox}

\end{document}